# Implementing Behavior Trees using Three-Valued Logic

Thibaud de Souza
1k1
Tainan, Taiwan
tea.desouza@gmail.com

*Abstract*—**With consideration to behavior trees and their relevance to planning and control, within and without game development, the distinction between stateful and stateless models is discussed; a three-valued logic bridging traditional control flow with behavior trees is introduced, and a C# implementation is presented.**

## I. INTRODUCTION

Introduced by Isla in 2008 [1], behavior trees (BT) constitute a control strategy for managing successful, failing and running tasks [2]; BT has applications in games, AI and robotics [3].

In video game design, BT solutions often assume a separation between technical design and engineering; visual solutions enable technical designers to interactively craft behavior trees; APIs are provided for engineers to integrate native tasks and custom decorators.

In mainstream languages such as Java, C++, C# or Python, behavior trees are often consumed via embedded, domain specific languages (EDSLs).

Both visual solutions and EDSLs have disadvantages. Programmers do not favor visual solutions, which often fall short with regard to integration with existing toolchains (testing, search, analysis). Implementing a comprehensive EDSL is a significant effort (on par with building a scripting runtime) and the result is not native of the host language, hindering readability and adding to both debugging and training costs.

Perhaps owing to these limitations, BT may have had a lesser impact on software engineering in general; this despite significant potential in clarifying and enhancing the asynchronous case logic inherent to many software applications.

The author author describe a novel approach combining native integration with good performance.

## II. BEHAVIOR TREES

As described in [2], BT embodies a task oriented approach to control. Assuming discrete time steps, each task is modeled as *failing*, *succeeding* or *complete*. Tasks are composed into hierarchies of *sequences* and *selectors*, where:

- A sequence iterates children until a failing task is encountered.

- A selector (fallback) iterates children until a successful task is encountered.

In Game AI and robotics, BT's field of applications overlaps with stronger planning strategies, such as goal oriented action planning (GOAP) [4] and hierarchical task networks (HTN) [5]; Isla touted BT's predictability as an advantage, since both players and game designers favor predictable designs [1]; BT generalizes Finite State Machines (FSMs) and decisions trees [6].

BT fits the wider framework of discrete control; in this respect, their appeal may extend beyond gaming and robotics, with potential applications to a range of software engineering problems.

BT occupies a middle ground between planning and traditional control flow. Conventional programming scales poorly to game AI problems, while STRIPS-inspired [7] approaches such as GOAP and HTN do cover wider problem domains, but with matching overheads (computational, production).

## III. STATELESS MODELS

With relevance to BT as a planning strategy, the distinction between stateful and stateless models is often overlooked; in order to motivate a stateless model, this distinction must be clarified.

Iovino et al. have detailed an algorithm for BT sequences, adapted here for comprehension [3]:

```
Sequence:
    for each task:
        Evaluate the task
        if the task is running:
            return running
        else if the task is failing:
            return failing
    return success
```

In this model a child task may reiterate even after it has completed or failed, and the details of what "evaluating" entails are left to specific task implementations.

This approach does not require storing information about the state of a task. Indeed, Iovino et al. emphasize that reiterating tasks ensures *reactiveness*, while briefly addressing how this may exacerbate the dithering problem described in [1].

In common implementations, tasks and composites are often stateful. Champandard and Dunstan's model [8] exemplifies this approach:

- Each task is an object or data structure.

- An init, update, end lifecycle is enforced via a formal interface.

- The return state of a task is stored between iterations, so that the init method is called once and, after a task

has completed or failed, later invocations of the same do not reiterate the task, yet return the previously stored state.

To understand how the divergence between stateful and stateless models may be affecting the design of behavior trees, we will use a simple example: "brewing coffee", which in a first approximation is modeled as a *sequence*:

| Making Coffee (stateful sequence) |
| --- |
| 1. Pour water in the kettle. |
| 2. Turn the kettle on. |
| 3. Pour ground coffee in the coffee pot. |
| 4. Wait for the kettle to turn off. |
| 5. Pour hot water in the pot. |
| 6. Wait a minute or two. |
| 7. Pour coffee in the cup. |

With a stateful model, tasks execute in order, with each task running once until complete or failed. If a task fails in the process of running the above sequence, the sequence itself has failed and the resulting state is stored in the sequence node. Once a stateful sequence has failed (or succeeded), a reset signal must be issued in order to reinitialize and re-run the sequence.

With a purely stateless model, ordering is not warranted and running tasks may be interrupted; thus the above sequence will proceed differently. It may be seen that, depending on how each task is implemented, steps 2-3-4 may cycle indefinitely.

Minding stateless execution, we then rewrite the above as a *selector*:

| Making Coffee (stateless selector) |
| --- |
| 1. If coffee is ready, pour into the cup. |
| 2. Wait until the coffee has infused. |
| 3. If the kettle is off, and the water is hot, pour water from the kettle to the coffee pot. |
| 4. Pour ground coffee into the coffee pot. |
| 5. If there is cold water in the kettle, turn the kettle on. |
| 7. Pour water from the water tap to the kettle. |

Since a selector implements a fallback strategy, later steps only evaluate if earlier steps are failing.

Comparing the two solutions above, we find that modeling a composite task as a stateful sequence is simpler since keeping track of the task index avoids re-checking prior steps; the resulting procedure, however, is reliant on a specific starting state (kettle empty, coffee pot empty, and so forth); further, it is also not designed to handle external interference (such as another person attempting to use the kettle).

Drawing information from world state at each iteration, the stateless approach biases design towards greater *resilience*. In the chosen example, it may be seen that the goal (hot coffee in a cup) will be attained regardless of the initial state, and external interface is better accounted for.

Duly noting that stateful approaches provide remedies in the form additional configuration and/or decorators (and the author's library also supports both stateless and stateful composites), it may be seen that a stateless approach focuses on desired outcomes, whereas a stateful approach, to a point, confuses the means for the end.

Still, where gaming applications are concerned, stateful composites are often useful because many steps are performative in nature (they do not result in a state change)

Our conclusion here is that, while availing both stateless and stateful composites is necessary, focusing on stateless models promotes correctness; from a production point of view, this encourages developers to address issues upfront instead of introducing bugs.

In the wider context of software engineering, tasks ordinarily produce outcomes; then, the stateless model may prevail.

## IV. ACTIVE LOGIC

Modern computing is grounded into Boolean logic, which informs the design of control statements, operators and loops; including the conditional statement (*if-else*), and the ternary operator ($x \, ? \, y : z$).

Conventionally, computer programs are designed as hierarchies of functions and sub-functions. Superficially this is similar to behavior trees. In BT, however, the assumption is that a sequence (or selector) may fail (or postpone) *at every single step*; when simulating (or in robotics: interacting with) real-time, dynamic environments, uncertainty and timeliness must be accounted for.

In contrast with binary logics, three-valued logics, first introduced by Peirce [9] and Post [10], are modeled as involving *true* (T), *false* (F) and *uncertain* (U) values and were developed to model reasoning under uncertainty; in modern computing a better known application may be SQL's 3VL [11], which leverages a subset of Kleene's logic [12] to help manage *null* database entries.

Similar to classical logic, three-valued logics are primarily concerned with immutable truths (be they static in general, or unchanging within the bounds of a reasoning context).

The author then propose an active logic, which purports selecting between available actions. Given the state of a task x, this logic will help us determine whether a composed task y should execute; with respect to discrete control, *indetermination* maps the *running* state; then, where time based applications are considered, uncertainty arises and resolves in time.

The logical conjunction and disjunction are modeled using the following truth tables:

| x ∧ y | Logical AND | | |
| --- | --- | --- | --- |
| | *F* | *U* | *T* |
| F | F | F | F |
| U | U | U | U |
| T | F | U | T |

| x ∨ y | Logical OR | | |
| --- | --- | --- | --- |
| | *F* | *U* | *T* |
| F | F | U | T |
| U | U | U | U |
| T | T | T | T |

Disregarding the handling of undetermined values, the above follows classic binary logic; superficially, it may be seen that compared with Kleene's logic [12], operators are similarly defined. However for both conjunctions and disjunctions, if the first operand is undefined, the second operand is ignored, and the output is always undetermined; the conjunction and disjunction are associative; they are not, (unlike in Kleene's logic) commutative.

In addition to the conjunction and disjunction, the presented calculus also uses lenient and strict combinators. Where either of two actions (which may execute simultaneously) realize the same intent, the lenient combinator is used:

| x + y | Lenient Combinator | | |
|---|---|---|---|
| | *F* | *U* | *T* |
| F | F | U | T |
| U | U | U | T |
| T | T | T | T |

Where two actions (which again, may execute simultaneously) must complete in order to achieve a designated composed action, the strict combinator is used.

| x * y | Strict Combinator | | |
|---|---|---|---|
| | *F* | *U* | *T* |
| F | F | F | F |
| U | F | U | U |
| T | F | U | T |

The strict/lenient combinators closely resemble Priest's min/max operators [13]; they are associative and commutative.

Using associativity we may verify that active logic propositions of the form $T \rightarrow t^1 \odot ... \odot t^n$ ($T$: a composite task, $t^1 ... t^n$: child tasks, $\odot$: the conjunction, disjunction, lenient or strict operator) map the *sequence* ($\wedge$), *selector* ($\vee$) and *parallel* (+, *) execution strategies.

## V. IMPLEMENTATION

The author have implemented active logic using the C# programming language. C# is an established, high level, general purpose, cross-platform object oriented language which has come to prominence in game development, with well regarded game engines using the language as their main scripting platform in lieu of interpreted languages such as Lua or Python.

The mapping between common BT node types and the present implementation is detailed in the following table.

| Node type | Active Logic implementation |
|---|---|
| Sequence | && |
| Selector (fallback) | \|\| |
| Parallel | * / + |
| Action | Native (any status expression) |
| Condition | Native (implicit conversion from bool) |

In this implementation, *status* is a *readonly struct* representing the state of a given task; status takes on values of -1 (*failing*), 0 (*running*) or 1 (*complete*); in some languages status may be represented as an enumeration; C#, however, does not associate operators with *enums*.

C# has limited, yet strong support for three-valued logic via (indirectly) overloading the short-circuiting operators && and ||. Leveraging this feature, a *sequence* may be written as…

        t0 && … && tn;

…while a *selector* is written as:

        t0 || … || tn;

Unlike many other languages (including C, C++, Rust, Python, …), overloading in C# preserves the short-circuiting behavior of the logical conditional operators; then, the sequences and selectors thus composed may use any expression returning, or convertible to, the *status* type.

In pseudo-code, the logical AND may be described as follows:

```
Operator && (x, y):
    evaluate x
    if x is not complete:
        return the state of x
    otherwise:
        evaluate y
        return the state of y
```

The logical OR is described as:

```
Operator || (x, y):
    evaluate x
    if x is not failing:
        return the state of x
    otherwise:
        evaluate y
        return the state of y
```

In addition to sequences and selectors, the lenient (+), strict (*) and disregard (%) combinators are used to model parallel execution.

The disregard operator (%) runs tasks in parallel. In this case, $x \% y \rightarrow x$, thus the return state of the second task is ignored. While this operator may be considered optional, it is useful in composing status expressions.

Finally, the negation, promotion, demotion and condone unary operators (!, +, -, ~) map common BT decorators.

| x | Unary operators | | | |
|---|---|---|---|---|
| | *!x* | *+x* | *-x* | *~x* |
| F | T | U | F | T |
| U | U | T | F | U |
| T | F | T | U | T |

With regard to performance, the discussed implementation stores behavior trees in program memory; status objects are stack-allocated, so that the memory overhead per instance is zero.

A reference implementation is available online [14].

## VI. SAMPLE PROGRAM

An example will illustrate what active logic programs look like; the following is taken from a simulation modeling wildlife and hunting-gathering.

```csharp
public class HunterGatherer : Actor {

    public Transform home;
    bool            indoors;
    HunterGathererAP π = new HunterGathererAP();

    override public status Step()
    => GameTime.isDay ? HuntAndGather()
                      : Rest();

    status HuntAndGather()
    => Exit(home, ref indoors)
       && Hunt() || Gather() || Roam();

    status Hunt()
    => π.game && Strike(prey.game);

    status Gather()
    => π.food && Pack(π.food);

    status Rest()
    => Enter(home, ref indoors) && Sleep();

    status Sleep() => Play("sleep");

}
```

In the above example the behavior tree is rooted in the *Step* function, however this may (without modification) be reused in a more complex model; the decomposition into functions and sub-functions is optional, yet improves clarity and maintainability.

π represents an apperception model used to capture actionable states in the environment; separating perception and apperception from control is good practice.

*Actor* defines low level actions common to many agents (such as locomotion or gesturing); this approach is suitable for small projects (or delegation should be preferred).

The use of expression bodied members (=>) is optional; status functions may include multiple statements and allow mixing traditional control statements with BT-styled sequence and selectors. The below example illustrates this.

```csharp
protected status Reach(Transform x,
                       float dist=1f,
                       float speed){

    if(transform.Has(x)) return done;
    if(!LookAt(x).complete) return cont;
    float δ = transform.MoveNear(x, dist,
                                 speed);

    return (δ == 0f)
           || Play(LocomotionAnim(speed));
}
```

## VII. APPLICATIONS AND BENEFITS

The presented approach results in concise, readable, memory efficient and less error prone BT programs. The effort needed to implement a comprehensive BT solution is reduced.

Status expressions may appear anywhere in a C# program. They seamlessly combine with existing facilities, so that a programmer may leverage features of the host language.

Beyond robotics and game AI, these benefits may extend the field of BT applications.

In designing and implementing user interfaces, uncertainty resolved in time is relevant to both user interaction (where the user are presented with linear or non linear choices resolved via asynchronous input) and cosmetic effects and animations.

More generally, the presented approach is compatible with problems involving either realtime or step-wise, iterative control. Problems involving realtime control include asynchronous communication (such as with computer networks), while iterative control is relevant to many algorithms and procedures.

## VIII. RELATED WORK AND CONCLUSION

In bridging the gap between binary logic and time based applications, this research expands the benefits of behavior trees and their range of application; the presented approach is compatible with many high level, functional and object oriented languages languages (including C++, Java, Rust and Swift).

Where non-game applications are considered, the distinction between stateful and stateless control appears especially relevant since the stateless model combines resilience and predictability.

This work partakes a focused effort to engineer robust, dependable behavior trees solutions. As such the *Active Logic* library [15] comprehends other constructs (such as stateful and mutable composites), along with specialized types for increased type safety and clarity (*certainties*), decorators, and extended tracing facilities (such as history recording).

Current work focuses on solutions to the dithering problem, game related applications (real-time, turn based) and physically based animation; in connection with utility AI [16], scalar status representations are also being investigated.

Present and current efforts have consistently pointed at significant constraints in how contemporary, broad purpose, high-level languages handle control; coordinated efforts may help relax these limitations.

Libraries and examples are made available on Github [14, 15].